\documentclass[11pt,a4paper]{article}

\usepackage[utf8]{inputenc}
\usepackage[T1]{fontenc}
\usepackage{xcolor}
\usepackage{amsmath,amssymb}
\usepackage{graphicx}
\usepackage{hyperref}
\usepackage{geometry}

\usepackage{cite}
\newcommand{\ignore}[1]{}

\title{\textbf{Why Pure Reasoning is Not Enough: Nature as the Source of Mathematical Innovation}
}
\author{Charanjit S. Jutla\\ \small \textit{IBM T. J. Watson Research Center } %\\ \small \texttt{email@example.com}
\\ \small \textit{Yorktown Heights, NY 10598}
\and Vimal Sharma \\
\small \textit{Independent} \\
\small \textit{Melbourne, FL 32940}
}
\date{\today}

\begin{document}
\maketitle

\begin{abstract}
We advance the hypothesis that human mathematical reasoning,
 constrained by both the undecidability and the computational intractability of even modest logical fragments,
relies fundamentally on pattern matching from domains external to pure deduction. The most prolific reservoir of such patterns is the natural world, whose physical laws and biological systems have undergone billions of years of ``pre‑computation'' and already exhibit surprisingly innovative solutions.  

To ground this claim, we trace the history of the Fourier transform and relevant mathematics, from the vibrating string controversy to the heat equation and subsequent formalisms prevalent in mathematics.
At each critical juncture, a physical problem forced the acceptance or creation of a mathematical tool that pure formal reasoning  failed to anticipate or, worse, human reasoning had resisted.

We further survey the landscape of logical complexity, from NP‑hard propositional satisfiability to the non‑elementary decision procedures for monadic second‑order theories, to demonstrate that even when a logic is decidable, the resources required for worst-case deduction are astronomically prohibitive.  We argue that these barriers make physics‑inspired pattern matching not just a historical accident but a cognitive necessity.  Finally, we draw the consequence for artificial intelligence: if pure reasoning is constitutively insufficient, then any system aiming at human‑level mathematical creativity must embed a vast store of cross‑domain patterns rather than rely on deduction alone.  This furnishes a principled justification for the enormous scale of contemporary large language models, and locates their present frontier at the boundary between recombining known patterns and extending the conceptual vocabulary itself.
\end{abstract}

%=====================================================================
\section{Introduction: Mathematics by Analogy}
\label{sec:intro}

Formal mathematical systems are limited by two profound facts: the existence of undecidable propositions (Gödel, Church, Turing) and the inherent incompleteness of any sufficiently powerful axiomatic framework \footnote{See Section~\ref{sec:undecide} for details of undecidability and seemingly simple fragments of logic that have non-elementary time complexity.}.  A purely deductive ``search'' for new theorems cannot systematically produce all true statements, nor can it efficiently navigate the vast space of possible propositions.  Yet mathematics progresses with astonishing speed, often by importing structures from the physical world.  We propose that the human mind copes with undecidability by using \emph{pattern matching}: recognising in a natural or physical system a solution that can be abstracted and formalised.  Nature, with its eons of evolutionary optimization or possibly due to physical selection (see for instance, the strong anthropic principle~\cite{weinberg,schombert,susskind2003}), has already explored and ``solved'' countless problems in ways that are highly non‑intuitive and effective.  The history of science is replete with examples where physics provided the germ of a mathematical innovation that pure logic alone would have been hard‑pressed to generate.

The present paper develops this thesis through a single, exceptionally rich case: the Fourier transform and its progeny.  We follow its path from the musical perception of harmony, through the mathematical scandal of representing an arbitrary function as a sum of sinusoids.  At every step, the catalyst was a concrete physical question, not an abstract mathematical puzzle.

After presenting the historical evidence, we situate our thesis within the existing literature, showing that while many authors have remarked on the physics‑first pattern or on the embodied nature of mathematics, the specific link to the undecidability barrier and the resultant \emph{cognitive necessity} of pattern matching from nature has not been explicitly argued.  We then draw a direct corollary for artificial intelligence: because pure reasoning is insufficient, any system aiming at human‑level mathematical and algorithmic creativity must embed vast stores of cross‑domain patterns, providing a fundamental justification for the enormous scale of contemporary large language models.  We conclude by outlining the broader implications of this view as well as addressing the computational complexity of pattern matching for theorem proving.

Taken together, the evidence assembled in this paper supports a single, sharp conclusion: pure deductive reasoning is not merely slow but \emph{constitutively insufficient} as an engine of mathematical discovery.  The undecidability of first‑order logic and the non‑elementary cost of even decidable fragments (Section~\ref{sec:undecide}) foreclose brute‑force search as a viable route to new mathematics, while the recurring historical pattern---from the vibrating string to the heat equation to the vector‑space concept itself---shows that the decisive innovations were read off from nature and only afterwards reconstructed as theorems.  The negative case of axiomatic set theory (Section~\ref{subsec:set-theory}) confirms the same law from the other side: mathematics that severs its physical grounding pays with foundational crises and scientific sterility.  Pattern matching from the physical world is therefore not an incidental heuristic but the mechanism by which mathematical cognition circumvents the barriers that pure logic cannot cross---and it is precisely this mechanism that any artificial system aspiring to comparable creativity must reproduce at scale.

\subsection{Before Fourier: The Vibrating String and Physical Overtones}
\label{sec:string}

The idea that a complex sound can be built from pure tones is an immediate biological experience.  The human ear performs a real‑time Fourier analysis, and music itself is an exploration of frequency ratios.  In the 1740s, this physical reality became a mathematical battlefield.

D'Alembert \cite{dalembert1747} derived the one‑dimensional wave equation and showed that the shape of a plucked string could be represented as the sum of two travelling waves (see Appendix~\ref{app:d-alembert} for details).  Daniel Bernoulli \cite{bernoulli1753} went further: he proposed that \emph{any} initial shape of the string could be expanded as an infinite sum of simple harmonic modes,
\[
y(x,0) = \sum_{n=1}^{\infty} a_n \sin\left( \frac{n\pi x}{L} \right),
\]
just as one hears a superposition of a fundamental and its overtones.  This was a purely physical hypothesis, grounded in music and mechanics.

The mathematical establishment of the time, led by Euler \cite{euler1749} and Lagrange \cite{lagrange1759}, rejected Bernoulli's claim (see~\cite{ravetz1961,kline1972} for details of this historical episode).  They argued that only ``analytic'' functions, those given by a single closed‑form expression, could be represented by a trigonometric series.  A triangular pluck was not analytic; therefore, they believed, Bernoulli's expansion could not hold.  The dispute was not resolved by logic but by the stubborn fact that a plucked string \emph{does} sound like a sum of sinusoidal components\footnote{That nature is replete with periodic phenomenon is obvious, but the formulation of sine and cosine as periodic functions were a direct consequence of observing planetary motions, with the current form of sine and cosine usually attributed to Aryabhata (500 CE)~\cite{sine-history}.}.  Physics pointed to a truth that the prevailing mathematical ontology could not accommodate.

\subsection{Orthogonality and the Emergence of Vector Spaces: Multiple Gifts from Physics}
\label{sec:orthogonal}

In the vibrating string debate, an equally profound property was hiding in plain sight. The sine functions that Bernoulli used are \emph{orthogonal} on the interval \([0,L]\):
\[
\int_0^L \sin\left(\frac{n\pi x}{L}\right) \sin\left(\frac{m\pi x}{L}\right) dx = 0 \quad \text{for } n \neq m.
\]
To a modern mathematician, this is a statement about an inner product on a function space. To an 18th‑century physicist, it was a consequence of the fact that different normal modes of a string do not exchange energy; they vibrate independently. The orthogonality was \emph{observed} in nature before it was abstracted into algebra.

\ignore{
When Fourier later applied the same basis to the heat equation \cite{fourier1822}, he exploited this orthogonality explicitly to compute the coefficients \(a_n\) and \(b_n\). Multiplying the series by \(\sin(m\pi x/L)\) and integrating term‑by‑term, all cross‑terms vanish, yielding an explicit formula for each coefficient. This procedure was not a theorem of pure linear algebra; it was a formalisation of the physical fact that heat diffusion separates into independent spatial modes.
}

\subsection*{The deeper root: vector spaces before Grassmann}

The abstract concept of a vector space — a set closed under addition and scalar multiplication — is often credited to Grassmann's Ausdehnungslehre (1844)~\cite{grassmann1844}, but its essential ingredients were already in widespread use, drawn directly from mechanics and geometry. The parallelogram law for composing forces appears in pseudo-Aristotelian mechanics, is made rigorous by Stevin (1586) and Newton (1687), and becomes standard for engineers throughout the 18th century. Descartes' analytic geometry (1637), the Argand plane for geometric representation of complex numbers (Wessel 1799, Argand 1806), and Hamilton's quaternions (1843) each contributed a physically motivated 
calculus of directed magnitudes.

Grassmann’s monumental contribution was to abstract these scattered, physically motivated calculi into a single general theory of vector spaces of arbitrary dimension. But his starting point was explicitly geometric and physical: he sought a theory of extension that could handle forces, currents, and crystallographic symmetries, topics on which his father had worked \cite{crowe1967}. Thus, the vector space concept itself—like the orthogonality that later gave it an inner product—emerged from the pattern matching of concrete physical and geometric situations, not from a purely deductive exploration of algebraic axioms.

\subsection{Fourier's Heat: The Theorem Forged in a Furnace}
\label{sec:heat}

As mentioned above, Bernoulli had already proposed that the motion of a vibrating string could be written as a superposition of simple harmonic modes,  
however, he gave no systematic method for determining the coefficients \(a_n\); his insight remained qualitative, grounded in the physical experience of hearing overtones, and he did not assert the representation for discontinuous functions.  
\ignore{Fourier, driven by the practical need to compute the evolution of \emph{any} initial temperature profile in the heat equation, invented the integral formulas for the coefficients by exploiting the orthogonality of the sine and cosine functions, the first explicit ``Fourier coefficients.''  
}

The decisive breakthrough came from an entirely different physical domain.  In 1807 Joseph Fourier submitted a memoir on the propagation of heat in solid bodies \cite{fourier1822}.  To solve the heat equation, he separated variables and obtained ordinary differential equations whose solutions were sines and cosines.  The general solution forced him to assert that \emph{any} initial temperature distribution \(f(x)\) could be written as
\[
f(x) = \sum_{n=0}^{\infty} \bigl( a_n \cos(nx) + b_n \sin(nx) \bigr),
\]
with the coefficients given by the now‑famous integrals. See Appendix~\ref{app:fourier-heat} for more details.

Fourier's claim was revolutionary: it included functions with jumps, corners, and other ``unnatural'' features.  Lagrange, who was present at the 1807 presentation, again objected on grounds of rigour.  Fourier's reply was quintessentially physical, the heat equation predicted a smooth evolution that matched experiment, and the trigonometric representation was a necessary consequence of the physics.  The \emph{Analytical Theory of Heat} (1822) became the cornerstone of Fourier analysis, not because the mathematics was complete, but because nature had spoken.

The statement that \emph{every} function, including discontinuous, non‑differentiable, and later even pathological ones, could be decomposed into perfectly smooth basis functions was so radical that it eventually required a complete rewriting of the concepts of function, integral, and convergence (see Appendix~\ref{app:beyond-periodic} for more pedagogical statements).

\subsection{Mathematical Rigour Follows the Physics}
\label{sec:rigour}

Once the physical legitimacy of the Fourier series was established, mathematicians were forced to build a formal framework that could justify it.  This process, far from being a triumph of autonomous logic, was a prolonged effort of \emph{catching up with nature}:

\begin{itemize}
    \item \textbf{Dirichlet (1829)} \cite{dirichlet1829} provided the first sufficient conditions for convergence of Fourier series (piecewise monotonic, finite number of jumps).  His work was directly motivated by the need to give precise meaning to Fourier's claims.
    \item \textbf{Riemann} (1854) \cite{riemann1854} developed his integral precisely to handle more general Fourier coefficients.  In his Habilitation thesis, he investigated the representability of functions by trigonometric series, and his integral was born from the difficulties of defining the Fourier coefficient integrals.
    \item \textbf{Cantor} (1872) \cite{cantor1872} began studying the sets of points where a trigonometric series might fail to represent a function uniquely; this research ultimately led him to invent set theory.
    \item \textbf{Lebesgue} (1902) \cite{lebesgue1902} introduced his measure‑theoretic integral and the \(L^2\) space framework, largely motivated by the convergence problems of Fourier series and the emerging needs of physics.
    \item \textbf{Schwartz} (1950) \cite{schwartz1950} created the theory of distributions to give a rigorous home to the Dirac delta function, a ``function'' that physicists (and Heaviside before them) had been using with great success in electromagnetism and quantum mechanics.  The delta function is the Fourier transform of a pure tone; once again, physical utility forced mathematical extension.
\end{itemize}

In every instance, the physical tool was in active use long before the mathematicians found a logically airtight home for it.  The direction of discovery was not from axioms to theorem, but from nature to formalisation.

\section{The Undecidability Connection}
\label{sec:undecide}

The history we have sketched is not merely a collection of anecdotes.  It illustrates a deep epistemic principle grounded in the logical limits of formal systems.

\subsection{Undecidability and Incompleteness as Real Barriers}
\label{sec:complexity}

Gödel’s incompleteness theorems (1931) demonstrated that any consistent, sufficiently expressive formal system contains true statements that cannot be proved within the system.  Church \cite{church1936} and Turing \cite{turing1936} later showed that even the decision problem for first‑order logic is undecidable: no algorithm can determine, for an arbitrary statement, whether it is logically valid.  These results are not just philosophical curiosities; they place hard, practical constraints on automated deduction.  A theorem prover operating solely on axioms and inference rules cannot, in general, decide the truth of arbitrary mathematical conjectures, and the search space for proofs grows hyper‑exponentially even in areas that are decidable.

\subsection*{Decidable but intractable: a hierarchy of computational hardness}

The undecidability of general predicate logic is only the most extreme manifestation of a deeper practical problem: even among \emph{decidable} fragments, the computational resources required to determine truth can be so enormous that purely deductive algorithms become useless for all but the tiniest problems.  A brief tour of the classical complexity hierarchy for logical decision problems makes this painfully clear.

\paragraph{Propositional logic: the base of the tower.}
The simplest non‑trivial logic is classical propositional calculus.  The problem of deciding whether a given propositional formula is a tautology is \textbf{co‑NP‑complete}, while satisfiability is \textbf{NP‑complete} \cite{cook1971}.  
\begin{quote}
\emph{Example.}  The formula
\[
((P \to Q) \land (Q \to R)) \to (P \to R)
\]
is a tautology; verifying such formulae for an input with hundreds of Boolean variables, however, may require exponential time in the worst case (unless P = NP).
\end{quote}
Already at this first rung, pure deduction encounters the exponential explosion that limits fully automatic reasoning for large formulas.

\paragraph{Quantified Boolean formulas: PSPACE‑completeness.}
If we add universal and existential quantifiers over propositional variables, we obtain the language of \textbf{Quantified Boolean Formulas} (QBF).  Deciding the truth of a QBF sentence is \textbf{PSPACE‑complete}, believed to be far harder than NP‑complete problems because PSPACE is not known to be contained in NP \cite{papadimitriou1994}.  
\begin{quote}
\emph{Example.}  The sentence
\[
\forall x\,\exists y\,\forall z\,\exists w\; \bigl( (x \land y) \lor (\neg x \land z)  \lor (\neg z \land w)\bigr)
\]
is true, but finding a proof (a winning strategy for the existential player) resembles a game tree of exponential depth.
\end{quote}
QBF can encode two‑player games and many problems in automated planning; its PSPACE hardness shows that even when decidability is trivial, the proof search is a severe bottleneck.

\paragraph{Temporal logics: from PSPACE to EXPTIME.}
Temporal logics are used to specify the behaviour of reactive systems over linear or branching time.  
\begin{itemize}
    \item \textbf{Linear Temporal Logic (LTL)} is \textbf{PSPACE‑complete}.  A typical specification is  
    \[
    \mathbf{G}\, (\text{request} \to \mathbf{F}\, \text{grant}),
    \]
    meaning ``globally (at all times), a request is eventually followed by a grant.''  Although the model‑checking problem for a single finite trace is tractable, the \emph{validity} problem for arbitrary formulas can require exploring all possible infinite sequences, hence the PSPACE hardness.
    \item \textbf{Computation Tree Logic (CTL)} is \textbf{EXPTIME‑complete} \cite{emerson1990}.  A branching‑time property like  
    \[
    \mathbf{AG}\, (\text{request} \to \mathbf{AF}\, \text{grant})
    \]
    asserts that along \emph{every} possible future path, a request will eventually be satisfied (along all possible futures!).  The EXPTIME hardness stems from the need to check the formula over the full, exponentially large computation tree.
\end{itemize}
Both logics are decidable, yet practical automated reasoning tools (e.g., model checkers) must employ sophisticated heuristics, symbolic representations, and pattern‑like abstractions to avoid brute‑force deduction.

\paragraph{First‑order theories of graphs: PSPACE completeness and undecidability.}
Restricting to pure first‑order logic, even the language of graph theory already exhibits extreme complexity.
%Consider the \textbf{Rado graph}, the unique countable homogeneous graph.  Its first‑order theory—the set of all sentences true in this single infinite graph—is decidable but \textbf{PSPACE‑complete} \cite{rado}.  
\ignore{
  \begin{quote}
\emph{Example.}  The extension axiom ``for any two disjoint finite sets of vertices there exists a vertex adjacent to all of the first and none of the second'' is a typical sentence true of the Rado graph; deciding whether an arbitrary first‑order sentence holds in the Rado graph requires polynomial space in the worst case.
  \end{quote}
  }
The \emph{validity problem over all finite graphs}—determining whether a first‑order sentence holds in \emph{every} finite graph—is \textbf{undecidable} (Trahtenbrot's theorem).  Thus, even within the relatively weak framework of first‑order logic, the structural richness of graphs pushes deduction either into intractable complexity or outright impossibility.

\paragraph{S1S and S2S: the non‑elementary abyss.}
The monadic second‑order theories of one successor (S1S) and of two successors (S2S) push complexity to an extreme.  
\begin{itemize}
    \item \textbf{S1S} is the theory of natural numbers with the successor function, allowing quantification over both individual numbers and \emph{sets} of numbers (or equivalently, the second order theory of linear order ($\mathbb{N}, <$) ~\cite{buchi1962}).  It is decidable, but its decision procedure requires time that grows faster than any elementary function (i.e., non‑elementary) \cite{meyer1973}.  
    \begin{quote}
    \emph{Example.}  The well‑ordering principle ``every non‑empty set of natural numbers has a least element'' is expressible  in S1S.
    \end{quote}
    \item \textbf{S2S} extends this to infinite binary trees, again with set quantifiers.  Rabin proved it decidable, but the complexity is still non‑elementary \cite{rabin1969}.  
    \begin{quote}
    \emph{Example.}  König's infinity lemma, ``every infinite, finitely branching tree has an infinite path,'' is naturally expressed  in S2S.
    \end{quote}
\end{itemize}

\paragraph{Why are second‑order theories of linear order so hard?}
The root of the non‑elementary blow‑up lies in the extraordinary expressive power of monadic second‑order quantification over a linear order, even one as simple as the natural numbers with the usual ``less‑than'' relation.
In monadic second‑order logic one can quantify not only over elements but over arbitrary \emph{sets} of elements.  Over a linear order, a set quantifier essentially allows a formula to name a subset of the domain, which can encode an arbitrary binary sequence or, over a tree, an arbitrary path.  A single second‑order existential quantifier already requires the decision procedure to guess the set’s characteristic function, and the quantifier alternations are the real killer.

For \textbf{S1S}, the decision procedure translates a formula into a finite automaton on infinite words (B\"uchi automaton) \cite{buchi1962}.  Each time a set quantifier is eliminated, the automaton’s state space undergoes a \emph{powerset} construction, yielding an exponential blow‑up.  With $k$ alternating quantifiers, the automaton’s size becomes a stack of $k$ exponentials.  Hence the decision time is at least $k$-fold exponential, i.e., non‑elementary in the quantifier depth.  In effect, the logic can express arbitrarily deep inductive definitions—for instance, the well‑ordering principle or the existence of winning strategies in increasingly long games—and the automaton must keep track of all possibilities.

\ignore{
\textbf{S2S} extends this to the infinite binary tree, whose topology is far richer than a line.  Even without quantifier alternations, tree automata (Rabin automata) \cite{rabin1969} have a more complex complementation procedure that already involves an exponential.  With alternations, the same stacking of exponentials pushes the total complexity into non‑elementary territory even for \emph{fixed} formulas with just a few alternations.  Moreover, the binary tree can encode the full second‑order theory of one successor, so S2S subsumes S1S and adds still more expressive power.
}

Thus, the underlying reason is that a linear order (or a tree) provides enough combinatorial structure for monadic second‑order logic to encode extremely deep and nested inductive reasoning.  The automata‑theoretic decision procedures, while proving decidability, reveal that any brute‑force deductive exploration must traverse a stack of exponentials, making the task physically impossible for all but the smallest formulas.

\ignore{
\paragraph{Why are S1S and S2S so hard?}  
The root of the non‑elementary blow‑up lies in the interplay between set quantifiers and the automata‑theoretic decision procedures.  In S1S, a formula with alternating quantifiers over sets is translated into an automaton on infinite words; each quantifier alternation causes the automaton's state space to grow by an \emph{exponential} power set construction.  With $k$ alternations, the resulting automaton has a size given by a stack of exponentials of height proportional to $k$, yielding decision times that are at least $k$‑fold exponential.  S2S replaces infinite words by infinite trees, where the underlying automata are inherently more complex even without quantifier alternations, and the same stacking effect pushes the complexity into non‑elementary territory already for fixed formulas.  Effectively, these theories encode arbitrarily high levels of inductive definitions, making any brute‑force proof search impossible in the physical universe.
}

Thus, even within the decidable fragment, a purely deductive machine faces either exponential, doubly exponential, or non‑elementary time demands.  This immense computational landscape reinforces our main argument: the human mind does not—and cannot—navigate mathematics by brute deduction.  It must rely on the compressed, pre‑computed patterns that nature and physics provide.

%\ignore{
\subsection*{Nature as the oracle}

These barriers force a different strategy.  As we have seen with the Fourier transform, the crucial ideas—decomposing an arbitrary function into sinusoids, exploiting orthogonality, constructing the FFT—did not emerge from a systematic logical exploration of function spaces.  They were observed first in physical systems (vibrating strings, heat diffusion, seismic waves, medical imaging) and then formalised.  The natural world acts as an oracle: it provides concrete instances of solutions that are already “true” in the physical domain, which mathematicians then generalise and axiomatise.  In this way, pattern matching from nature circumvents the undecidability barrier.  A mathematician does not derive the Fourier series from first principles; she recognises it in the music of a string and then works backwards to a formal justification.
%}

\section{Negative Cases}
\subsection{Axiomatic Set Theory}
\label{subsec:set-theory}

The thesis is strengthened by examining what happens when mathematics attempts to advance without physical grounding. Cantor's set theory, as noted in Section~\ref{sec:rigour}, originated in the concrete context of Fourier series. However, the subsequent programme of Hilbert, Russell, and Zermelo–Fraenkel attempted to place all of mathematics on a purely axiomatic foundation, deliberately severing the connection to physical intuition.

The result was instructive. The naive formalism immediately produced paradoxes - Russell's paradox, the Burali-Forti paradox, the Richard paradox - that could not be resolved by pure deduction and required external heuristics such as the axiom of choice, the axiom of regularity, and Zermelo's axiom schema to patch. Each of these additions was motivated not by derivation from prior axioms but by informal judgements about what 'should' be true — a covert form of pattern matching from everyday intuition about collections and size.

Furthermore, the most celebrated results of pure set theory - Gödel's constructible universe L, the independence of the continuum hypothesis, the large cardinal hierarchy - have produced essentially no tools that practising scientists, engineers, or even most pure mathematicians use. Compare this to Fourier analysis, which underpins signal processing, quantum mechanics, medical imaging, and communications. The difference is precisely the presence or absence of physical grounding. Set theory, when it severed its connection to nature, became locally productive but globally sterile for the purposes of the broader scientific enterprise.

This negative case makes the thesis falsifiable and considerably stronger: it is not merely that physical inspiration happens to have been present at key moments, but that its absence predicts a characteristic failure mode - foundational crisis followed by patching with informal axioms, and disconnection from scientific practice.
\subsection{The Proof-Assistant Objection}

A natural objection runs as follows. Modern proof assistants, such as Lean, Coq, Isabelle, can verify and even discover non-trivial mathematical results by pure deduction from axioms, without any physical input. Does this not refute the thesis?

It does not, for a subtle but decisive reason. Proof assistants automate the verification of deductive steps; they do not supply the high-level structure of a proof. The lemma decomposition, the choice of intermediate goals, the selection of which definitions to unfold, all of these are provided by a human mathematician who has absorbed, through years of exposure, exactly the cross-domain patterns we describe. When a proof assistant 'discovers' a result, it is searching within a space whose architecture was designed by a human who brought physical and geometric intuition to bear. The assistant is an extremely fast checker, not an autonomous discoverer.
The recent success of AI-assisted theorem proving (e.g., AlphaProof) is fully consistent with the thesis: these systems work by learning from vast corpora of human-written mathematics, which already encodes the physically-grounded patterns accumulated over centuries. They are pattern stores, not deductive engines.

\subsection{Naive Counter-arguments to ``Physics First, Abstraction later''.}

It has been noted that Kepler obtained elliptical orbits for the planets, whereas the ellipse had already been formulated as an abstract object by Greek mathematicians (Apollonius, ca.~200~BCE) centuries earlier; thus, one may argue that  abstraction preceded the physics. However, this is counter to what actually happened. The conic sections were available for nearly two millennia, and the Greeks even possessed a serviceable observational model of planetary motion in the epicyclic (circular) constructions of Ptolemy; yet neither pure reasoning nor the pre-existing abstraction ever attached the ellipse to the heavens. Kepler himself began from the a priori conviction, inherited from the Greeks and Copernicus, that orbits \emph{must} be built from perfect circles. What forced him off that conviction was not deduction but Tycho Brahe's naked-eye observations of Mars: his best circular model fit the data to within eight minutes of arc. Yet, Kepler rejected it precisely because he trusted Tycho's measurements more than he trusted the circle, remarking that those eight arcminutes ``led the way to the reformation of all of astronomy'' \cite{keplerAN1609,voelkel2001}. Only after this, and after a laborious empirical search through ovals and other figures, did the ellipse (with the Sun at one focus) emerge as the shape the data demanded \cite{caspar1993}. His path was moreover guided by a \emph{physical} analogy—a motive force emanating from the Sun, conceived on the model of magnetism (after Gilbert~\cite{gilbert1600}) and of light, which yielded the area law before the ellipse itself \cite{stephenson1987}. Far from a case of abstraction preceding physics, Kepler is a paradigm of nature overriding pure reason: the abstract object sat idle for two thousand years, and only observation could pick it out and bind it to physical reality. (It was Newton, in 1687, who later \emph{derived} Kepler's ellipses from the inverse-square law: abstraction catching up with a physically established fact, exactly as our thesis predicts.)

Similarly, it is naively argued that Lobachevsky discovered non-Euclidean (hyperbolic) geometry before it found physical application in general relativity, so that here abstraction clearly preceded physics. Two points blunt this objection. First, the physical stimulus was not absent but diffuse: hyperbolic geometry arose from the two-thousand-year effort to prove Euclid's parallel postulate, and the postulate itself is an idealisation of physical space distilled from surveying, drawing, and astronomy. So, the very problem Lobachevsky attacked was rooted in the geometry of the physical world. Second, the negative-curvature geometry Lobachevsky formalised is precisely the one whose closest tangible model, the pseudosphere, was identified only later (by Beltrami~\cite{beltrami1868}), and whose decisive physical employment did indeed wait for relativity. We therefore do not press Lobachevsky as a positive example of ``physics first''; we note only that its physical antecedent (the parallel-postulate problem) is real, and that the case is the exception that the set-theory discussion of Section~\ref{subsec:set-theory} already leads us to expect - abstraction that outran its physical grounding paid for it with a long period of apparent sterility.

Representation theory and non-commutative algebra were likewise formalised before being gainfully employed in quantum mechanics. But the physical study of light had long before supplied the operative pattern: the composition of successive optical elements (refractions and translations along an axis) is a composition of linear \emph{ray-transfer} transformations, and Hamilton's optical–mechanical analogy (1830s) recast both optics and mechanics as a single calculus of transformations acting on a phase space \cite{hamilton1833}. Non-commutativity, the fact that the order of two operations matters, was thus a tangible feature of composing physical motions and optical maps long before it was axiomatised as an algebraic structure.

\subsection{Turing Machines and Recursive Function Theory}

The Turing machine formulation and Godel's recursive function theory  is not just a construct to formalize man-made computing machines, but algorithms in general. Note that since millennia,  it is known that humans can execute/calculate iteratively and algorithmically\footnote{The biological functions of animals and even cells can also be seen as computational/algorithmic, however this point of view is more widely accepted  only in recent times.}. 
%the first computing machine that mathematicians like Turing were aware of is the human brain, as it was already known that neurons in the brain led to cognition.
Having said that, there is reason to believe that there are human innovations that have no counterparts in nature, and possibly even no chain of inspirations. We address this in the next subsection~\ref{subsec:gears}.

\subsection{When the Artefact Has No Direct Natural Precedent}
\label{subsec:gears}

A natural objection to the thesis that mathematical innovation is
driven by pattern‑matching from the physical world is the existence
of human‑made constructs that seem to have no obvious counterpart in
nature.  Mechanical gears, for instance, are used in automata to
replicate sequential labour, yet interlocking toothed wheels do not
appear in the natural environment in the same direct way that
vibrating strings or diffusing heat do.  At first glance, such
artefacts might suggest that human creativity can operate without
borrowing from the physical world.

However, a closer look reveals that the pattern‑matching underlying
these inventions operates at the level of \emph{function} and
\emph{motivation} rather than at the level of the finished artefact’s
surface form.  The wheel itself likely emerged from the observation of
rolling logs or round fruits; the lever from the experience of lifting objects with a
rigid bar.  The gear can be seen as a recombination of these two
nature‑derived primitives---rotation and leverage---driven by the
functional goal of transmitting and transforming force, a goal that
itself is inspired by the observation that biological organisms
(muscles, limbs) already solve analogous problems.  The motivation to
build automata comes directly from watching animals and humans perform
repetitive labour; the mechanical solution does not copy the form of a
limb, but it borrows the \emph{functional pattern} of sequential,
periodic motion and force amplification.

This distinction is precisely the one we draw between
\emph{vocabulary recombination} and \emph{vocabulary extension}.
Gears, like many mechanical inventions, are a recombination of
existing concepts (wheel, lever, wedge) that already had clear
physical antecedents.  The invention is creative, but it stays within
the conceptual space defined by nature‑inspired primitives.  In
contrast, the truly radical mathematical abstractions discussed in
this paper---measure theory, transfinite set theory---required
\emph{vocabulary extension}, the introduction of genuinely new
primitives.  Even then, the new primitive did not arise in a vacuum:
Lebesgue’s measure abstracted the physical intuition of length, and
Cantor’s transfinite numbers generalised the intuitive distinction
between finite and infinite.

Thus, the apparent counterexample of artefacts without natural
precedent actually strengthens the thesis.  It shows that
nature‑inspired pattern‑matching can be sufficiently abstract to
operate at the level of function and need, and that the resulting
innovation can take a form not directly observed in the physical
world, as long as it is built by recombining concepts whose origin
is in physical experience.  This also explains why current LLMs,
which excel at vocabulary recombination, can propose mechanical
designs or analogical transfers within a known conceptual space,
but struggle with the kind of vocabulary extension that produced
measure theory or set theory: the latter requires a new primitive
whose genesis is not merely a recombination, and the oracle of
nature in its raw form may be the only source of such primitives.
%=====================================================================
\section{Conclusion}
\label{sec:conclusion}

We have argued that mathematical innovation is not, at its core, a
process of deduction from axioms, but one of \emph{pattern matching}
from the physical world.  The argument rests on two pillars that
reinforce one another.  The first is a limitative fact about formal
systems: general first‑order validity is undecidable, and even the
decidable fragments surveyed in Section~\ref{sec:undecide}---from
NP‑complete propositional satisfiability up to the non‑elementary
decision procedures for the monadic second‑order theories S1S and
S2S---demand computational resources so vast that brute‑force
deductive search is not a physically realisable strategy for
discovering new mathematics.  The second is historical: traced in
detail through the Fourier transform and its progeny
(Section~\ref{sec:intro}), the decisive conceptual leaps---the
representation of an arbitrary function as a sum of sinusoids, the
orthogonality that underlies inner‑product spaces, the vector‑space
abstraction itself, and ultimately the passage to the Fourier
integral---were each read off from a concrete physical situation and
only afterwards reconstructed as rigorous theorems.  In every case the
direction of discovery ran from nature to formalisation, not the
reverse.

The thesis is not merely descriptive but falsifiable, and the negative
cases sharpen rather than weaken it.  Axiomatic set theory
(Section~\ref{subsec:set-theory}), having deliberately severed its ties
to physical intuition, exhibited exactly the failure mode the thesis
predicts: foundational paradoxes patched by informally motivated
axioms, and a striking disconnection from the working needs of science.
The apparent counterexamples---Kepler's ellipse, Lobachevsky's
hyperbolic geometry, non‑commutative algebra, and the abstract Turing
machine---dissolve on inspection into confirmations: in each the
physical stimulus was present, if sometimes diffuse, and abstraction
that outran its physical grounding paid for it with long periods of
apparent sterility.  Even artefacts with no direct natural precedent,
such as mechanical gears (Section~\ref{subsec:gears}), turn out to be
recombinations of nature‑derived primitives, driven by functional goals
that themselves come from observing the physical and biological world.

The most consequential corollary is for artificial intelligence.  If
pure reasoning is constitutively insufficient---if the engine of
mathematical creativity is the recognition and transfer of patterns
accumulated from an enormous range of physical and cross‑domain
experience---then a machine aspiring to human‑level mathematical and
algorithmic discovery cannot be built as a deductive engine alone.  It
must instead embody a vast store of such patterns.  This is precisely
what the scale of contemporary large language models provides, and it
furnishes a principled, rather than merely empirical, justification for
that scale: the size of the pattern store is not incidental to the
task but demanded by it.  At the same time, the distinction between
\emph{vocabulary recombination} and \emph{vocabulary extension}
(Section~\ref{subsec:gears}) marks the current frontier.  Present
systems excel at recombining concepts within a known space, but the
genuinely new primitives---measure, the transfinite, the distribution---
have historically required an external oracle, namely nature in its raw
form.  Whether an artificial system can be given access to such an
oracle, or must forever remain a recombiner of patterns it has been
shown, is the open question our thesis leaves for future work.

\section{Acknowledgment}

We used Deepseek and Opus Sonnet to provide us with historical references and help write some of the sections, especially the details in the appendix. The paper was also reviewed with Opus-4.8 which led to negative cases section, for which we then provided the counterarguments.
%=====================================================================

\appendix

\section{D'Alembert's Travelling Wave Solution for the Plucked String}
\label{app:d-alembert}

In his 1747 memoir \cite{dalembert1747}, Jean le Rond d'Alembert considered a string of length $L$ fixed at both ends, set into vibration by an initial displacement.  Let $y(x,t)$ denote the transverse displacement at position $x$ and time $t$.  By applying Newton's second law to an infinitesimal segment of the string and assuming small displacements, he arrived at the one‑dimensional wave equation
\begin{equation}
\frac{\partial^2 y}{\partial t^2} = c^2 \frac{\partial^2 y}{\partial x^2},
\label{eq:wave}
\end{equation}
where $c$ is a constant determined by the tension and linear density of the string.

D'Alembert observed that any function of the form $f(x+ct)$ or $g(x-ct)$, when substituted into (\ref{eq:wave}), satisfies the equation identically.  By linearity, the general solution is the sum of two such travelling waves,
\begin{equation}
y(x,t) = f(x+ct) + g(x-ct),
\label{eq:general}
\end{equation}
where $f$ and $g$ are arbitrary functions determined by the initial conditions.

For a plucked string released from rest, the initial conditions are
\[
y(x,0) = \phi(x), \qquad \frac{\partial y}{\partial t}(x,0) = 0,
\]
with $\phi(x)$ describing the initial shape of the string.  Substituting $t=0$ into (\ref{eq:general}) gives
\[
f(x) + g(x) = \phi(x).
\]
Differentiating (\ref{eq:general}) with respect to $t$ and setting $t=0$ yields
\[
c\,f'(x) - c\,g'(x) = 0 \quad\Longrightarrow\quad f'(x) = g'(x),
\]
so $f(x) = g(x) + \text{constant}$.  The fixed endpoints of the string eventually force the constant to be zero, and one obtains $f(x) = g(x) = \frac{1}{2}\phi(x)$ (after extending $\phi$ as an odd periodic function to satisfy the boundary conditions).  The full solution then becomes
\begin{equation}
y(x,t) = \frac{1}{2}\bigl[\phi(x+ct) + \phi(x-ct)\bigr].
\label{eq:d-alembert-solution}
\end{equation}

Equation (\ref{eq:d-alembert-solution}) shows that the subsequent motion is the superposition of two waves, each preserving the shape of the initial displacement but scaled by one‑half, travelling in opposite directions with speed $c$.  D'Alembert therefore demonstrated that \emph{any} initial shape $\phi(x)$—subject only to the physical boundary conditions—can be represented as the sum of two propagating waves.

This result was groundbreaking: it provided the first analytical solution of a partial differential equation governing a continuous mechanical system.  However, it also sparked the controversy with Euler, Bernoulli, and Lagrange.  D'Alembert, like Euler, believed that the initial function $\phi(x)$ had to be given by a single ``analytic'' expression—a smooth, twice‑differentiable formula—whereas Bernoulli argued, on physical grounds, that plucked strings with sharp corners (triangular shapes) should also be admissible.  The mathematical community at the time could not reconcile the abstract demand for analyticity with the physical reality of non‑differentiable initial data.  It was precisely this tension between pure logic and observed nature that foreshadowed the later vindication of Fourier's even bolder claim.
\section{Fourier's Derivation of the Heat Equation and Its Trigonometric Solution}
\label{app:fourier-heat}

Joseph Fourier began his investigation of heat diffusion around 1804, motivated by practical problems of temperature distribution in solids.  His complete theory appeared in the \emph{Théorie analytique de la chaleur} (1822) \cite{fourier1822}.  The following outlines his physical reasoning and the mathematical leaps that gave birth to the Fourier series.

\subsection*{The heat equation from physical principles}

Consider a thin, homogeneous rod of length $L$ with insulated sides, so that heat flows only along the $x$‑direction.  Let $u(x,t)$ be the temperature at position $x$ and time $t$.  Fourier postulated two fundamental physical laws:

\begin{enumerate}
    \item \textbf{Newton's law of cooling} (local form): The rate of heat flow across a cross‑section is proportional to the temperature gradient.  Expressed with Fourier's notation, the heat flux is $-K \frac{\partial u}{\partial x}$, where $K>0$ is the thermal conductivity.
    \item \textbf{Conservation of energy}: The net heat influx into a small segment $[x, x+\Delta x]$ equals the rate of change of its internal energy, which is proportional to $\frac{\partial u}{\partial t}$ times the heat capacity.
\end{enumerate}

Balancing these effects for an infinitesimal element and taking $\Delta x \to 0$ leads to the one‑dimensional heat equation
\begin{equation}
\frac{\partial u}{\partial t} = \alpha \frac{\partial^2 u}{\partial x^2},
\label{eq:heat}
\end{equation}
where $\alpha = K/(\rho c)$ is the thermal diffusivity (a constant depending on material properties).

%\subsection*{Why the product ansatz? Historical and physical motivations}

\ignore{
Beyond mathematical precedent, a simple physical observation supported the product form.  If the initial temperature profile of the rod were a pure sine wave $\sin(\pi x/L)$, one would expect it to decay exponentially over time while retaining its sinusoidal shape.  That is, the rod should cool as a single, independent ``mode.''  The product $X(x)T(t)$ captures exactly this behaviour: the spatial configuration remains fixed, while the amplitude diminishes according to its own decay constant.  For an arbitrary initial profile, the principle of superposition then suggests that the evolution is a sum of such independent modes.

Finally, the product ansatz is a powerful mathematical expedient: it reduces the partial differential equation to two ordinary differential equations, which were far easier to solve with the 18th‑century analytical methods.  Fourier's true originality, therefore, lay not in the separation of variables itself, but in his audacious claim that \emph{any} function could be built from an infinite sum of these elementary product solutions—and in his invention of the coefficient formulas that extracted the weights from the initial data.
}%ignore

\subsection*{Separation of variables and the eigenvalue problem}

Fourier sought solutions of (\ref{eq:heat}) on the interval $0 \le x \le L$ with fixed boundary conditions, for instance $u(0,t)=u(L,t)=0$ (the ends held at zero temperature).  He assumed that the solution could be written as a product of a spatial and a temporal part\footnote{Fourier's crucial step—assuming the solution could be written as a product $u(x,t)=X(x)T(t)$—was not a leap into the unknown.  Rather, it was a direct import of a technique already proven successful in the study of vibrating strings.  Daniel Bernoulli (1753) had shown that the motion of a string could be decomposed into normal modes of the form $\sin(n\pi x/L)\cos(n\pi ct/L)$, each a product of a spatial shape and a temporal oscillation.  The method of separation of variables had become part of the analytical toolkit for linear partial differential equations with constant coefficients.
},
\[
u(x,t) = X(x)\, T(t).
\]
Substituting into (\ref{eq:heat}) and dividing by $X(x)T(t)$ gives
\[
\frac{T'(t)}{\alpha T(t)} = \frac{X''(x)}{X(x)} = -\lambda,
\]
where $\lambda$ is a separation constant (the negative sign is chosen because physical intuition suggests that temperature decays to an equilibrium).  This yields two ordinary differential equations:
\begin{align}
X''(x) + \lambda X(x) &= 0, \label{eq:spatial} \\
T'(t) + \alpha \lambda T(t) &= 0. \label{eq:temporal}
\end{align}

The boundary conditions $u(0,t)=u(L,t)=0$ force $X(0)=X(L)=0$.  Equation (\ref{eq:spatial}) with these homogeneous Dirichlet conditions is a Sturm–Liouville eigenvalue problem.  Its non‑trivial solutions exist only for
\[
\lambda = \lambda_n = \left(\frac{n\pi}{L}\right)^2, \qquad n = 1,2,3,\dots
\]
and the corresponding eigenfunctions are the sine functions
\[
X_n(x) = \sin\left(\frac{n\pi x}{L}\right).
\]

For each $n$, the temporal equation (\ref{eq:temporal}) gives
\[
T_n(t) = e^{-\alpha (n\pi/L)^2 t}.
\]

\subsection*{The general solution and the trigonometric series}

Since the heat equation is linear, any linear combination of the separated solutions is also a solution.  Fourier therefore proposed that the most general solution satisfying the boundary conditions could be written as an infinite series:
\begin{equation}
u(x,t) = \sum_{n=1}^{\infty} b_n \, e^{-\alpha (n\pi/L)^2 t} \, \sin\left(\frac{n\pi x}{L}\right).
\label{eq:series}
\end{equation}

To match an \emph{arbitrary} initial temperature distribution $u(x,0) = f(x)$, the coefficients $b_n$ must satisfy
\[
f(x) = \sum_{n=1}^{\infty} b_n \sin\left(\frac{n\pi x}{L}\right).
\]
Fourier now exploited a crucial physical fact which we already discussed in Section~\ref{sec:orthogonal}: the sine functions are orthogonal on $[0,L]$.
\ignore{
\[
\int_0^L \sin\left(\frac{n\pi x}{L}\right) \sin\left(\frac{m\pi x}{L}\right) dx = 
\begin{cases}
0, & n \neq m, \\
L/2, & n = m.
\end{cases}
\]
}
Multiplying the series by $\sin(m\pi x/L)$, integrating term‑by‑term from $0$ to $L$, and assuming that term‑wise integration is valid, all cross‑terms vanish and one obtains an explicit formula for each coefficient:
\begin{equation}
b_n = \frac{2}{L} \int_0^L f(x) \sin\left(\frac{n\pi x}{L}\right) dx.
\label{eq:fourier-coeff}
\end{equation}

\subsection*{The full trigonometric series: general boundary conditions and periodic functions}

Fourier did not stop with the sine series.  He also considered rods with insulated ends (Neumann boundary conditions, $\frac{\partial u}{\partial x}=0$ at $x=0,L$), which lead to a cosine series:
\[
f(x) = \frac{a_0}{2} + \sum_{n=1}^{\infty} a_n \cos\left(\frac{n\pi x}{L}\right),
\qquad
a_n = \frac{2}{L} \int_0^L f(x) \cos\left(\frac{n\pi x}{L}\right) dx.
\]

More fundamentally, Fourier recognized that an arbitrary function defined on a symmetric interval $[-L,L]$ and extended periodically can be represented by a combination of sines and cosines, the full Fourier series:
\begin{equation}
f(x) = \frac{a_0}{2} + \sum_{n=1}^{\infty} \left[ a_n \cos\left(\frac{n\pi x}{L}\right) + b_n \sin\left(\frac{n\pi x}{L}\right) \right],
\label{eq:full-series}
\end{equation}
with coefficients
\[
a_n = \frac{1}{L} \int_{-L}^{L} f(x) \cos\left(\frac{n\pi x}{L}\right) dx, \qquad
b_n = \frac{1}{L} \int_{-L}^{L} f(x) \sin\left(\frac{n\pi x}{L}\right) dx.
\]

\ignore{This is the form that appears in the main paper’s statement: Fourier asserted that \emph{any} function could be expressed as a sum of sines and cosines.  The sine‑only and cosine‑only series are special cases that arise when the function possesses odd or even symmetry, or when specific boundary conditions dictate which terms survive.
}

\subsection*{The bold claim and its legacy}

What set Fourier apart from his predecessors was his insistence that the representation (\ref{eq:series})–(\ref{eq:fourier-coeff}) holds for \emph{any} function $f(x)$ that is physically admissible as an initial temperature profile—even those with jumps, corners, or pieces defined by different analytic expressions.  This was a radical departure from 18th‑century notions of a function, which required a single closed‑form expression.  Lagrange and others objected fiercely, but Fourier's reply was that the physical success of the heat equation in predicting actual temperature evolutions vindicated the mathematics.

Fourier's method not only solved the heat problem but also introduced the general technique of expanding arbitrary functions in orthogonal eigenfunctions of a differential operator.  The integrals (\ref{eq:fourier-coeff}) are the first instance of what we now call Fourier coefficients, and the series (\ref{eq:series}) is the first explicit Fourier sine series.  It was this concrete, physically motivated construction that forced mathematicians to generalise the concepts of function, convergence, and integration, ultimately reshaping the entire landscape of analysis.

\section{Beyond the Finite Interval: Periodicity, Decay, and the Fourier Integral}
\label{app:beyond-periodic}

The sine and cosine series that Fourier employed to solve the heat equation on a finite rod are, by construction, periodic functions.  When one writes
\[
f(x) = \sum_{n=1}^{\infty} b_n \sin\left(\frac{n\pi x}{L}\right) \qquad (0 \le x \le L),
\]
the right‑hand side, viewed as a function on the whole real line, is odd and periodic with period $2L$.  Consequently, the equality holds only for $x \in [0,L]$; outside this interval the series reproduces the periodic extension of $f$, not the original function (unless $f$ itself already satisfies those symmetry and periodicity conditions).  This is a fundamental limitation: a trigonometric series cannot represent a function that is genuinely non‑periodic over the entire real line, nor can it represent a function defined on an infinite domain without imposing an artificial periodicity.

\subsection*{Functions that do not vanish at infinity}

A second, related limitation appears when one attempts to represent a function $f(x)$ defined on the whole real line $(-\infty,\infty)$ by a trigonometric series of the form $\sum a_n e^{inx}$.  Such a series is either a function on the circle (if $x$ is taken modulo $2\pi$) or, if interpreted on $\mathbb{R}$, it converges to a $2\pi$‑periodic function.  It cannot represent a non‑periodic function, and in particular it cannot represent a function that does not decay to zero as $|x| \to \infty$ unless one allows for a continuous spectrum of frequencies.

Fourier himself recognised this difficulty.  In the final chapters of the \emph{Théorie analytique de la chaleur} \cite{fourier1822}, he considered the problem of heat diffusion in an infinite solid, where no natural finite interval exists.  He argued heuristically that the discrete sum over integers $n$ should pass to an integral over a continuous frequency variable.  Formally, if one takes a function defined on $[-L,L]$ and lets $L \to \infty$, the Fourier series
\[
f(x) = \sum_{k=-\infty}^{\infty} c_k e^{i k \pi x/L}, \qquad
c_k = \frac{1}{2L} \int_{-L}^{L} f(\xi) e^{-i k \pi \xi/L} \, d\xi,
\]
transforms, under appropriate convergence assumptions, into the Fourier integral representation
\begin{equation}
f(x) = \int_{-\infty}^{\infty} \widehat{f}(\omega) \, e^{i\omega x} \, \frac{d\omega}{2\pi}, \qquad
\widehat{f}(\omega) = \int_{-\infty}^{\infty} f(x) \, e^{-i\omega x} \, dx.
\label{eq:fourier-integral}
\end{equation}

Here $\widehat{f}(\omega)$ is the Fourier transform of $f$.  This representation holds for a wide class of absolutely integrable functions ($f \in L^1(\mathbb{R})$), and later extensions to square‑integrable functions (Plancherel's theorem) and tempered distributions (Schwartz \cite{schwartz1950}) made it applicable to physically important objects such as the Dirac delta function, which does not decay at infinity and is not an ordinary function.

\subsection*{The periodic versus non‑periodic dichotomy}

The dichotomy between Fourier series and the Fourier integral is a manifestation of the underlying group structure: periodic functions live on the circle, which is a compact abelian group, while functions on the line live on a non‑compact abelian group.  The spectral decomposition of a function on a compact group yields a discrete sum (a series), whereas the decomposition on a non‑compact group requires a continuous integral.  This distinction was only fully clarified in the 20th century with the development of abstract harmonic analysis (Weil, Pontryagin).

\subsection*{Significance for the main thesis}

The historical transition from the Fourier series to the Fourier integral is itself another example of physics driving mathematical innovation.  Fourier's need to solve the heat equation on infinite domains forced him to invent what we now call the Fourier transform, a tool whose full justification took over a century and motivated the creation of the Lebesgue integral, the theory of distributions, and ultimately harmonic analysis on locally compact groups.  The pattern is the same: a concrete physical problem—the cooling of an unbounded solid—exposed the limitations of the existing mathematical machinery, and mathematicians followed, formalising the continuous spectral decomposition that nature had already “used” in the propagation of heat.

\end{document}